\documentclass[twoside,11pt]{article}
\usepackage{jmlr2e}
\usepackage{booktabs}
\usepackage{amsmath}
\usepackage{multirow}
\usepackage{color, colortbl}
\definecolor{Gray1}{gray}{0.7}
\definecolor{Gray2}{gray}{0.9}

\usepackage{lastpage}
\jmlrheading{}{2024}{1-\pageref{LastPage}}{1/21; Revised 5/22}{9/22}{21-0000}{Sonit Singh}
\ShortHeadings{Clinical Context-aware Radiology Report Generation}{Sonit Singh}
\firstpageno{1}

\begin{document}

\title{Clinical Context-aware Radiology Report Generation from Medical Images using Transformers}

\author{\name Sonit Singh \email sonit.singh@unsw.edu.au \\
       \addr School of Computer Science and Engineering\\
       Faculty of Engineering, University of New South Wales\\
       Kensington, NSW 2052, Australia
       \AND
       %\name Author Two \email two@cs.berkeley.edu \\
       \addr School of Computing\\
       Faculty of Science and Engineering, Macquarie University\\
       Macquarie Park, NSW 2109, Australia}

\editor{My editor}

\maketitle

\begin{abstract}%   <- trailing '%' for backward compatibility of .sty file
 Recent developments in the field of Natural Language Processing, especially language models such as the transformer have brought state-of-the-art results in language understanding and language generation. In this work, we investigate the use of the transformer model for radiology report generation from chest X-rays. We also highlight limitations in evaluating radiology report generation using only the standard language generation metrics. We then applied a transformer based radiology report generation architecture, and also compare the performance of a transformer based decoder with the recurrence based decoder. Experiments were performed using the IU-CXR dataset, showing superior results to its LSTM counterpart and being significantly faster. Finally, we identify the need of evaluating radiology report generation system using both language generation metrics and classification metrics, which helps to provide robust measure of generated reports in terms of their coherence and diagnostic value. 
\end{abstract}


\section{Introduction}

With advancements in Computer Vision (CV) and Natural Language Processing (NLP) techniques, researchers started exploring tasks at their intersection~\citep{Vinyals:2015:NIC,Singh:2018:pushing}. On the CV side, improved Convolutional Neural Network (CNN) architectures~\citep{He:Zhang:2016:Deep_residual_learning} have improved results for various CV tasks such as image classification, object detection, and image segmentation. On the NLP side, advanced language models such as Transformers~\citep{Vaswani:2017:attention_is_all_you_need} have resulted in a more accurate natural language understanding and text generation. Taking inspiration from the machine translation research~\citep{Sutskever:2014:Seq2Seq}, the task of image captioning aims at generating natural language description of content present in an image. It requires an algorithm to understand and model the relationship between visual and textual elements, and to generate a sequence of output words. Analogous to image captioning, radiology report generation aims at generating a textual radiology report describing key findings in the medical image. Automated radiology report generation from medical images is incredibly challenging as the model needs to have an understanding of the various medical conditions, grounding medical conditions in the image, and generating a coherent and long text outlining normal findings and abnormalities. There has been research towards automatic report generation~\citep{jing:2018:on_the_automatic,Xue:2018:multimodal_recurrent_model,Singh:2019:from_chest_X-rays,Singh:2021:show_tell_summarise}, adapting methods from generic image captioning, and are based on \emph{encoder-decoder} framework. In this framework, CNN is an encoder, which extracts semantic information in the image, whereas a RNN is the decoder, which decodes the extracted image features into a text sequence. The vanilla RNN models have limitations in terms of vanishing and exploding gradients. To overcome this, the LSTM and the GRU networks are preferred. However, RNNs have the complex addressing and overwriting mechanisms combined with inherently sequential processing problems~\citep{Zhu:Li:2018:captioning_transformer}. The training time of an RNN also increases with an increase in the sequence length, which inherently influences the training in parallel. 

\begin{figure}
    \centering
    \includegraphics[scale=0.45]{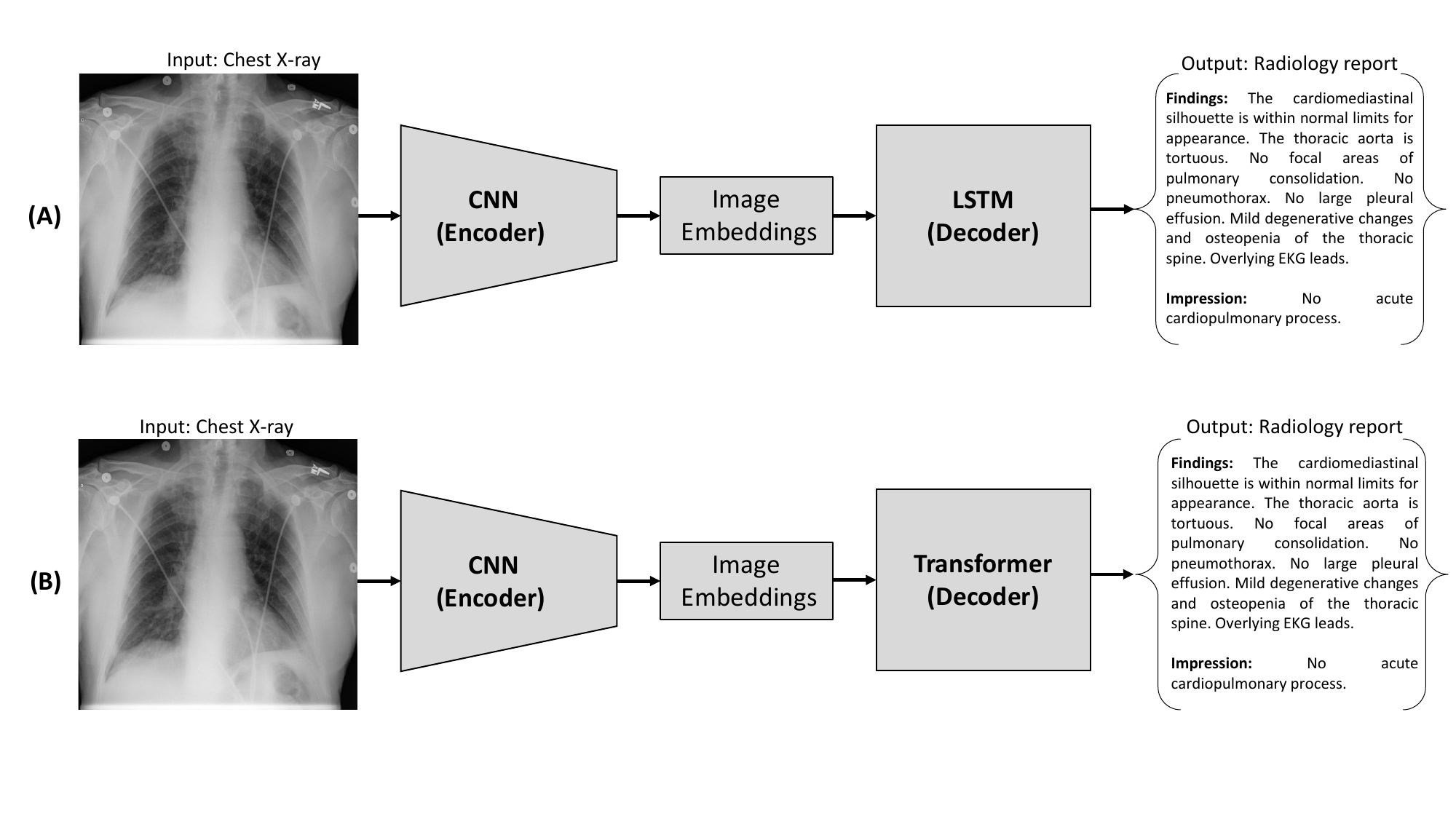}
    \caption{Block diagram of the (A) CNN + LSTM; and (B) CNN + Transformer model. In the proposed model (B), the CNN encoder extracts chest X-ray embeddings and the Transformer model (decoder) generates corresponding radiology report.}
    \label{fig:block_diagram}
\end{figure}

Recent developments in NLP, namely the Transformer architecture~\cite{Vaswani:2017:attention_is_all_you_need} have led to significant performance improvements for various tasks such as machine translation, text generation, and language understanding. The transformer model contains stacked attention mechanism, which can draw global dependencies between input and output as well as helping the model to attend to salient regions in the image. Also, transformers proved to be faster to train and can leverage GPU parallelisation, overcoming the sequential nature of recurrent models. Figure~\ref{fig:block_diagram} shows block diagram of generation radiology reports using LSTM or Transformer as a decoder. This motivates us to investigate the use of transformer model for radiology report generation. In this study, we use a pre-trained CNN model as an encoder to get visual features and use a transformer as a decoder to generate radiology reports from chest X-rays. We make use of publicly available dataset with pairs of chest X-rays and textual radiology reports, namely, the Indiana University Chest X-Ray collection (IU-CXR) dataset~\cite{Demner-Fushman:2015:Preparating_a_collection}. Experimental results in terms of standard natural language generation metrics showed that the transformer model provides competitive results compared to standard RNN model, yet being much faster and can be trained by parallel optimisation. The advantages of the proposed methodology are as follows: First, the simplicity of the transformer framework relieves the limitations of RNNs. Second, the application of a self-attention mechanism in the decoder helps to remember long-range dependencies which are helpful in generating radiology reports which are often lengthy text. Third, the transformer framework provides faster and parallel execution, in turn helping in faster radiology report generation which is important to speed up clinical workflow, improving patient care, and stratification of cases that need urgent attention. 

In this chapter, we make the following contributions:
\begin{itemize}
    \item We propose a transformer model based radiology report generation from the chest X-ray, which replaces the recurrent architecture with the self-attention mechanism. 
    
    \item We conduct extensive experiments by tuning various parameters to our validate hypothesis that transformer model provides improvement in quality of radiology report generation. 
    
    \item We highlight the limitations of evaluating radiology report generation using only the standard language generation metrics and provide the need to have both the language generation and clinical context-aware classification metrics. 
\end{itemize}

%%%%%%%%%%%%%%%%%%%%%%%%%%%%%%%%%%%%%%%%%%%%%%%%%%%%%%%%%%%%%%%%%%%%%%%%%
%%%%%%%%%%%%%%%%%%%%%%%%%%%%%%%%%%%%%%%%%%%%%%%%%%%%%%%%%%%%%%%%%%%%%%%%%
\section{Related Work}
One of the early works towards automated radiology report generation is by~\citep{Kisilev:MRG:2015} where feature based approach is applied to predict categorical BI-RADS descriptors for breast lesions. Three main descriptors, namely, \emph{shape}, \emph{margin}, and \emph{density} are used to train a classifier. Further categories for each feature are: \textbf{shape}: \{\emph{oval}, \emph{round}, \emph{irregular}\}, \textbf{margin}: \{\emph{circumscribed},  \emph{indistinct}, \emph{spiculated}, \emph{microlobulated}, \emph{obscured}\}, and \textbf{density}: \{\emph{non\-homogeneous},  \emph{homogeneous}\}. Although, the trained model can identify these features given a input image and fill them into a fixed template, the approach is limited to keywords and do not support coherent and free-form radiology report. \cite{Shin:2016:Learning_to_read_Chest_X-rays} used a cascaded CNN-RNN captioning model to generate description about the detected disease. Their model could generate individual words, however the generated words are not coherent and can be difficult to comprehend. This is due to the poor information extraction and language modelling capabilities. Later \citep{Zhang:2017:MDNet} proposed a CNN-RNN model enhanced by an auxiliary attention sharpening (AAS) module to automatically generate medical imaging report. They also demonstrated the corresponding attention areas of image descriptions. Their proposed model can generate more natural sentences but the length of each sentence is limited to $59$ words. In addition, the content of a generated report is limited to five topics. \citep{jing:2018:on_the_automatic} proposed a hierarchical co-attention based model for generating medical imaging report. Their proposed model have capability to attend image features and predict semantic tags while exploring joint effects of visual and semantic information. Also, their model can generate long sentence descriptions in reports by incorporating a hierarchical Long-Short Term Memory (LSTM) network. However, they concatenated findings and impression section into a single text for generation task. Also, the experimental results show that their model was prone to generating false positives because of the interference of irrelevant tags. 

\citep{Li:2018:hybrid_retrieval_generation} built a hierarchical reinforced agent, which introduced reinforcement learning and template based language generation method for medical image report generation. Their agent effectively utilise the benefit of retrieval and generation based approaches. However, the heavy involvement of pre-processing in extracting templates makes the method heuristic and difficult to generalise to other datasets and applications. \cite{Xiong:2019:Reinforced_Transformer_MIC} proposed a novel hierarchical neural network based Reinforced Transformer for medical image captioning (RTMIC) to generate coherent informative medical imaging report. The transformer module speeds up the training process by enabling parallel computing in feed-forward layers. Also, using bottom-up attention using pre-trained DenseNet model on ChestX-ray14 dataset that leads to more accurate pathological terminologies in the generated sentences. The use of reinforcement learning based training method addresses the discrepancy between Maximum Likelihood Estimation (MLE) based training objective and evaluation metrics of interest. However, only findings section was considered in the study and impression section is not included.

\cite{Yin:2019:mic_HRNN} proposed a hierarchical recurrent neural network (HRNN) based medical report generation model. They also introduced a topic matching mechanism to HRNN, so as to make generated reports more accurate and diverse. The HRNN consists of two RNNs, a sentence RNN and a word RNN. The sentence RNN takes detected abnormalities and the features maps extracted by the CNN as inputs, and then generates several topic vectors. Given a topic vector, the word RNN produces an appropriate sentence. However, the underlying assumption that most sentences are only related to a single disease or part of the location of the medical image is wrong. Moreover, they concatenate findings and impression sections as single text and regard the combined text as the ground truth report. \cite{Zeng:2018:generating_ultrasound_descriptions} proposed a coarse-to-fine ultrasound image captioning ensemble model that can generate description of ultrasound images. First an organ classifier detect organ present in an ultrasound and then a respective encoder-decoder framework generates report for respective organ.

%%%%%%%%%%%%%%%%%%%%%%%%%%%%%%%%%%%%%%%%%%%%%%%%%%%%%%%%%%%%%%%%%%%%%%%%%
%%%%%%%%%%%%%%%%%%%%%%%%%%%%%%%%%%%%%%%%%%%%%%%%%%%%%%%%%%%%%%%%%%%%%%%%%
\section{Methodology}
In this section, we provide the problem definition, overview of the transformer model, which is the building block of our CNN + Transformer model for radiology report generation. We then provide an overview of encoder-decoder framework for radiology report generation where CNN is the encoder whereas either LSTM or Transformer is the decoder. 

\subsection{Problem definition}
Given we have image-text pair data for the radiology report generation task, we use supervised learning to train the proposed model. Given the image-text pair $(I,S)$, we train the model to minimise the cross-entropy loss as follows:
\begin{equation}
    \text{log}\  p(S/I) = \sum_{t=0}^{N} \text{log} \ p(S_t | S_0, S_1, \ldots, S_{t-1}; \theta)
\end{equation}
\noindent where $\theta$ are the parameters of the model, $I$ is the chest X-ray and $S$ is the ground-truth radiology report having $N$ tokens.

During inference, each token is generated sequentially. We input the start token $<\text{start}>$, and generate the first word with probability $p(y_1| \theta, I)$. Afterwards, we get the dictionary probability $y_1 \sim p(y_1|\theta, I)$ at the first iteration. We use the beam search method to select the best possible word out of possible cases. The generated word $y_1$ is fed back into the network to generate the second word, and so on. The process will be continued until we reach the $<end>$ token or we reach the maximum length of the generated radiology report. 

\subsection{Transformer model}
In this section, we briefly summarise the original transformer model from its original authors~\cite{Vaswani:2017:attention_is_all_you_need} and outline its adaptation for generating radiology reports from chest X-rays. 

\subsubsection{Attention in Transformer}
The transformer model~\citep{Vaswani:2017:attention_is_all_you_need} has now become the de-facto model for most of the NLP tasks. The transformer model was first applied to the machine translation task, achieving state-of-the-art results, which can be trained faster due to parallel optimisation with less training costs. Before this, recurrent neural networks (RNNs) were established as the state-of-the-art approach for most of the NLP tasks. The transformer model is solely based on the attention mechanism without relying on recurrence and convolutions. In RNNs, the hidden state of the current time step, $h_t$, depends on the previous hidden state, $h_{t-1}$, creating a sequential execution. To overcome the inherent recurrence in RNN model, the transformer reformulate the calculation of the hidden state, where the hidden state of current time step, $h_t$, only depends on the feature embeddings of the input image and history words, rather than the previous hidden state, $h_{t-1}$. This helps the transformer model to be executed in parallel. 

There are two attention mechanisms in the transformer model, namely, \emph{scaled dot-product attention} and \emph{multi-head attention}. The Transformer uses scaled dot-product attention to deduce the correlation between queries $Q$ and keys $K$ and then obtains the weighted sum of the values $V$. The scaled dot-product attention is the basis of multi-head attention, and is computed as follows:
\begin{equation}
    \text{Attention}(Q,K,V) = \text{softmax}(\frac{QK^T}{\sqrt{d_k}})V
\end{equation}

\noindent where the attention inputs are composed of queries $Q \in \mathcal{R}^{L \times d_k}$,  key $K \in \mathcal{R}^{L \times d_k}$, and value $V \in \mathcal{R}^{L \times d_v}$. The queries $Q$, keys $K$, and values $V$ have dimensions $d_k$, $d_k$, and $d_v$ respectively. $L$ is the sequence length. $\frac{1}{\sqrt{d_k}}$ is used to counteract the effect of the large values of $d_k$ pushing the softmax function into the regions of small gradients. The scaled dot-product attention is different from the conventional attention mechanism as its attention weights are computed by dot-product operation. The resulting dot-product attention $Attention(Q,K,V) \in \mathcal{R}^{L \times d_v}$ is faster and more space efficient in practice because it can be implemented by parallel optimisation. All intra-modality and cross-modality interactions between word and image-level features are modelled via the scaled dot-product attention. 

Compared with single attention, multi-head attention can learn different representation sub-spaces at different positions. It contains $h$ identical attention heads, and each head is a scaled dot-product attention, performing the attention function on queries, keys, and values independently. Finally, $h$ attention outputs are concatenated and projected back to the original dimension, producing the final values. For each of the head, the attention output is computed as follows:

\begin{equation}
    \text{head}_i = \text{Attention}(QW_i^Q, KW_i^K, VW_i^V)
\end{equation}

The multi-head attention concatenates the heads as:
\begin{equation}
    \text{Multihead}(Q,K,V) = \text{Concat}(\text{head}_1, \text{head}_2, \ldots, \text{head}_h)W^O
\end{equation}

\noindent where $W^O \in \mathcal{R}^{hd_v \times d_{model}}$, $W_i^Q \in \mathcal{R}^{d_{model} \times d_{k}}$, $W_i^K \in \mathcal{R}^{d_{model} \times d_{k}}$, $W_i^V \in \mathcal{R}^{d_{model} \times d_{v}}$ are trainable projection matrices. To reduce the overall computation cost, \cite{Vaswani:2017:attention_is_all_you_need} projected the original dimension of $d_{model} = 512$ onto $d_k = d_v = d_{model}/h = 64$ and $h=8$. Self-attention is a specific case of multi-head attention where the queries, keys, and values are all from the same hidden layer. 

\subsubsection{Feed-forward network}
The Transformer model has two-layered fully-connected feed-forward network. The feed-forward network takes the input from the multi-head attention layer, and further transforms it through two linear projections with ReLU activation function. These feed-forwards layers are applied to each position separately and can improve the non-linearity of the transformer model.

\begin{equation}
    \text{FFN}(x) = \text{max}(0, xW_1 + b_1)W_2 + b_2
\end{equation}

\noindent where $W_1  \in \mathcal{R}^{d_{model} \times d_{h}}$, $W_2  \in \mathcal{R}^{d_{h} \times d_{model}}$ are trainable projection matrices; $b_1 \in \mathcal{R}^{d_h}$ and $b_2 \in \mathcal{R}^{d_{model}}$ are the biases; and $d_h$ is the hidden state size.

\subsubsection{Positional encoding}
Given that transformer contains no recurrence or convolutions, the sequential order is not leveraged and can notably affect the quality of generated descriptions. To provide sequence order information, positional embeddings are added, which are defined as follows:
\begin{equation}
    \text{PE}_{(pos, 2i)} = sin(pos/10000^{2i/d_{model}})
\end{equation}

\begin{equation}
    \text{PE}_{(pos, 2i+1)} = cos(pos/10000^{2i/d_{model}})
\end{equation}

\noindent where $pos$ is the absolute position of the token in the text and $i$ denotes the corresponding dimension of embeddings. Since the multi-head attention and feed-forward network contains no convolutional layers or recurrent cells, positional encoding is essential for leveraging the relative position information in sequence. 

\subsection{Radiology report generation with the Transformer model}
\begin{figure}
    \centering
    \includegraphics[width=17cm, height=7cm]{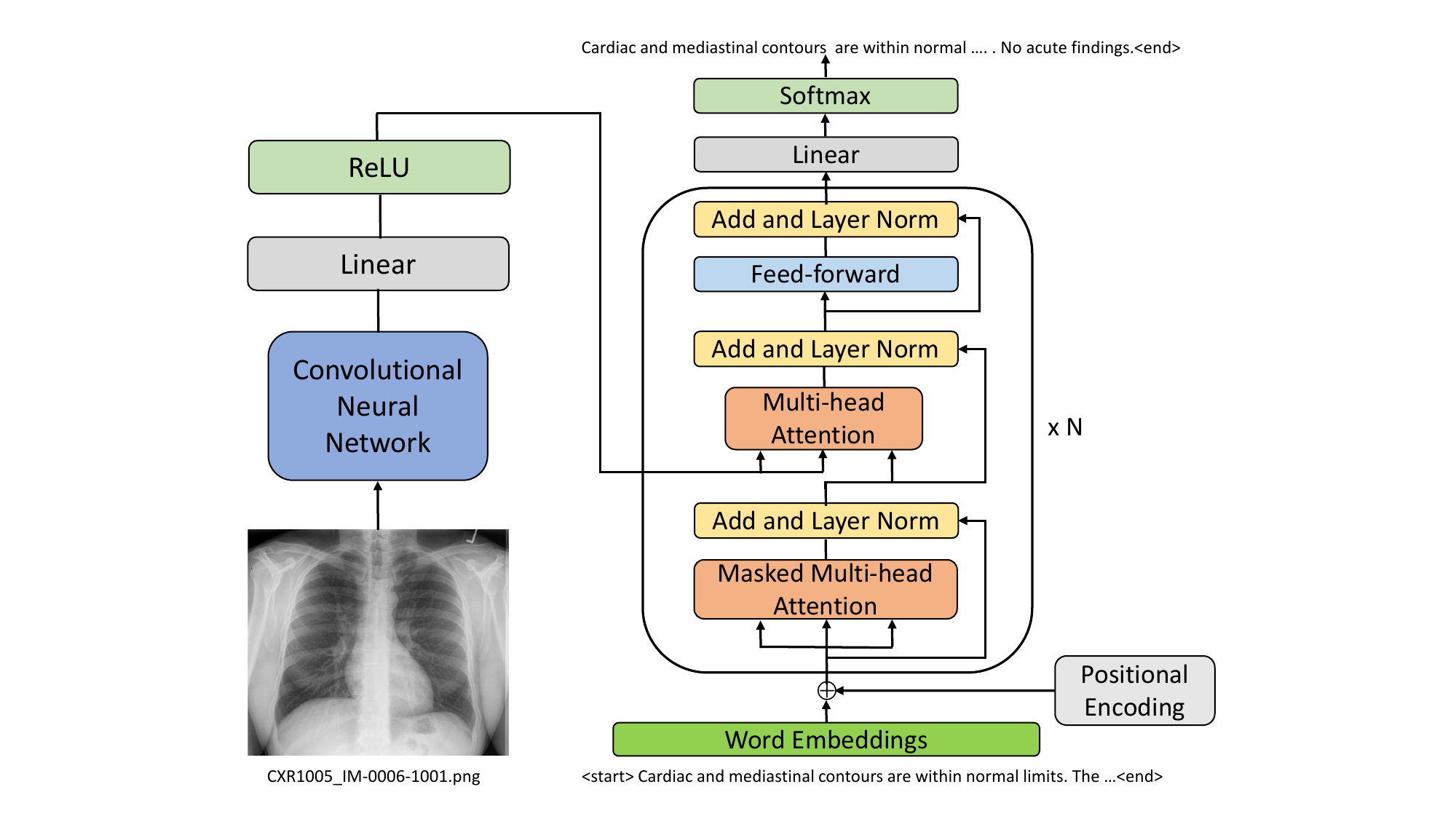}
    \caption{The encoder-decoder framework of the proposed CNN+Transformer model. The encoder is the convolutional neural network, such as the ResNet model. The decoder is the Transformer model. The Transformer decoder can have $N$ identical decoder layers. $<start>$ and $<end>$ tokens are added at the beginning and end of the radiology report text, respectively.}
    \label{fig:cnn+transformer}
\end{figure}

The proposed CNN+Transformer model is shown in Figure~\ref{fig:cnn+transformer}. It is based on the \emph{encoder-decoder} framework, where the encoder is a CNN and the decoder is the \emph{Transformer}. In order to check the effectiveness of the Transformer decoder, we also used Long-Short Term Memory (LSTM) network as the decoder, keeping CNN as an encoder. This helps to validate the effectiveness of the transformer model for radiology report generation.

\subsubsection{Encoder}
Different from the original transformer model, which is developed for NLP tasks, we use the CNN model as our encoder to extract semantic information of medical images. For the machine translation task for which the transformer model is originally proposed, the transformer model can encode the sequential input into a context vector, which then get decoded by the decoder to the output sequence. However, different from the machine translation task, the radiology report generation task has as input the medical image and as output the textual description of the image. In order to capture the semantic meaning of the input medical image, we use a state-of-the-art CNN, namely, the \emph{ResNet}~\cite{He:Zhang:2016:Deep_residual_learning} that are based on residual connections and its variants. 

\subsubsection{Decoder}
In order to set up a baseline to check the effectiveness of the transformer model, we use LSTM network as the baseline decoder for generating radiology report from medical images. We then, replace the LSTM decoder with the Transformer model, which relies on the stacked multi-head attention mechanism. The transformer model is made up of a stack of $N$ identical layers, where each layer has three sub-layers. The first uses the multi-head self-attention mechanism. This layer has a masked mechanism for preventing this model from seeing the future information. This masked mechanism can ensure the model generates the current word with only the previous words. The second layer is a multi-head attention layer without the masked mechanism. It performs the multi-head attention over the output of the first layer. This layer is the core layer to correlate the text information and the image information with the attention mechanism. The third layer is a simple, position-wise fully connected feed-forward network. The Transformer performs a residual connection around each of the three sub-layers, followed by layer normalisation. At the top, we add a fully connected layer and a softmax layer to project the output of the Transformer to the probabilities for the whole sentence. Different from the LSTM, all the words in the sentence can be generated in parallel. Given, the Transformer model follows an encoder-decoder architecture, we only use the decoder part for generating text but replace the default encoder with the CNN model as outlined in the previous section. 

\subsection{Limitations of NLG metrics}
\begin{figure}
    \centering
    \includegraphics[scale=0.45]{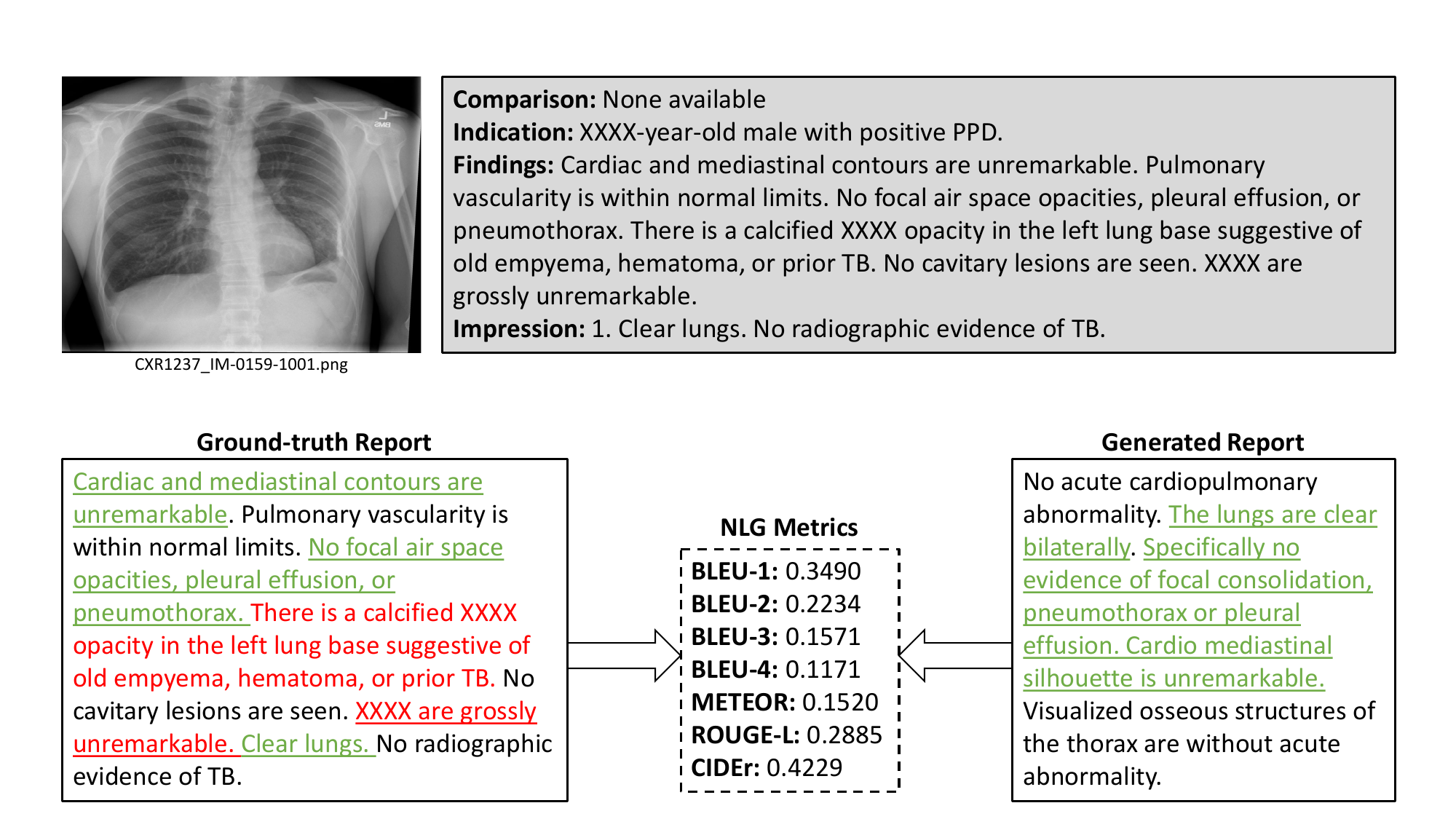}
    \caption{A selected sample case to highlight issues in evaluating radiology report with natural language generation (NLG) metrics.}
    \label{fig:evaluation_example_case}
\end{figure}

The performance of radiology report generation methods is mainly evaluated using natural language generation (NLG) metrics. The traditional NLG metrics such as the BLEU~\citep{papineni:2002:bleu} score do check word overlap in the reference report and the generated report, they do not provide information about the clinical value of the generated report. In Figure~\ref{fig:evaluation_example_case}, we highlight an example case, comparing the ground-truth radiology report with the generated report. Based on the NLG scores, a BLEU-1 score of $0.3490$ indicates that the model in generating radiology reports is comparable to the ground-truth report. However, the generated report misses some key findings of \emph{opacity} and \emph{empyema}. The NLG metrics are good at finding word overlap between the generated and ground-truth report, but completely lack clinical context, which is important from a diagnostic viewpoint. For instance, two sentence, namely, ``No focal areas of pulmonary consolidation" and ``Focal areas of pulmonary consolidation" are syntactically similar, but mean opposite in clinical terms. This highlights limitation of the NLG metrics in evaluating radiology reports with clinical context. 

Given that the NLG metrics do not specifically evaluate the clinical value of generated reports, there is the need to have complementary evaluation methods. To measure the diagnostic value of the generated reports, Keywords Accuracy (KA)~\cite{Xue:2018:multimodal_recurrent_model}, Diagnostic Content Score (DCS)~\cite{Babar:2021:evaluating_diagnostic_content}, and Clinical Coherence Score~\cite{Liu:2019:Clinically_accurate_CXR_generation} have been proposed. The KA metric calculates the ratio of the number of clinical keywords in the generated report to the number of clinical keywords in the ground-truth report. The KA metric has the ability to measure the overlap of clinical terms in generated and reference reports. However, to apply the KA metric, we need to first create a keyword dictionary from the Medical Text Indexer (MTI) annotations, which is a manual, time-consuming, and tedious task. This limits the usability of the KA metric in evaluating clinical context in large-scale generated radiology reports. The diagnostic content score (DCS) metric is based on a probabilistic model that predicts class labels in the generated report. Although, the DCS metric can provide diagnostic metric, similar to standard classification metric, it has two main limitations. First, to train the probabilistic model, we need tags associated with each radiology report. Second, the performance of the trained probabilistic model is not going to be 100\% accurate, leaving room of errors in the evaluation. The clinical coherence reward, which is based on the CheXpert labeler, can compare thoracic observations in the generated report and the ground-truth report. The CheXpert labeler is non-differentiable, because of which it can't directly be used to calculate loss function comparing ground-truth observations with predicted observations from generated report during model training. \cite{lovelace:2020:Learning_to_generate} developed a differentiable approximation of the CheXpert labeler by utilising the Gumbel-Softmax method. Our proposed method of evaluating radiology reports from a clinical perspective is also based on the CheXpert labeler, however, we overcome the issue of the non-differentiable problem by utilising CheXpert at the classification label and the generated report level, which is described in the next section.

\subsection{Evaluating radiology report from clinical perspective}
\begin{figure}
    \centering
    \includegraphics[scale=0.45]{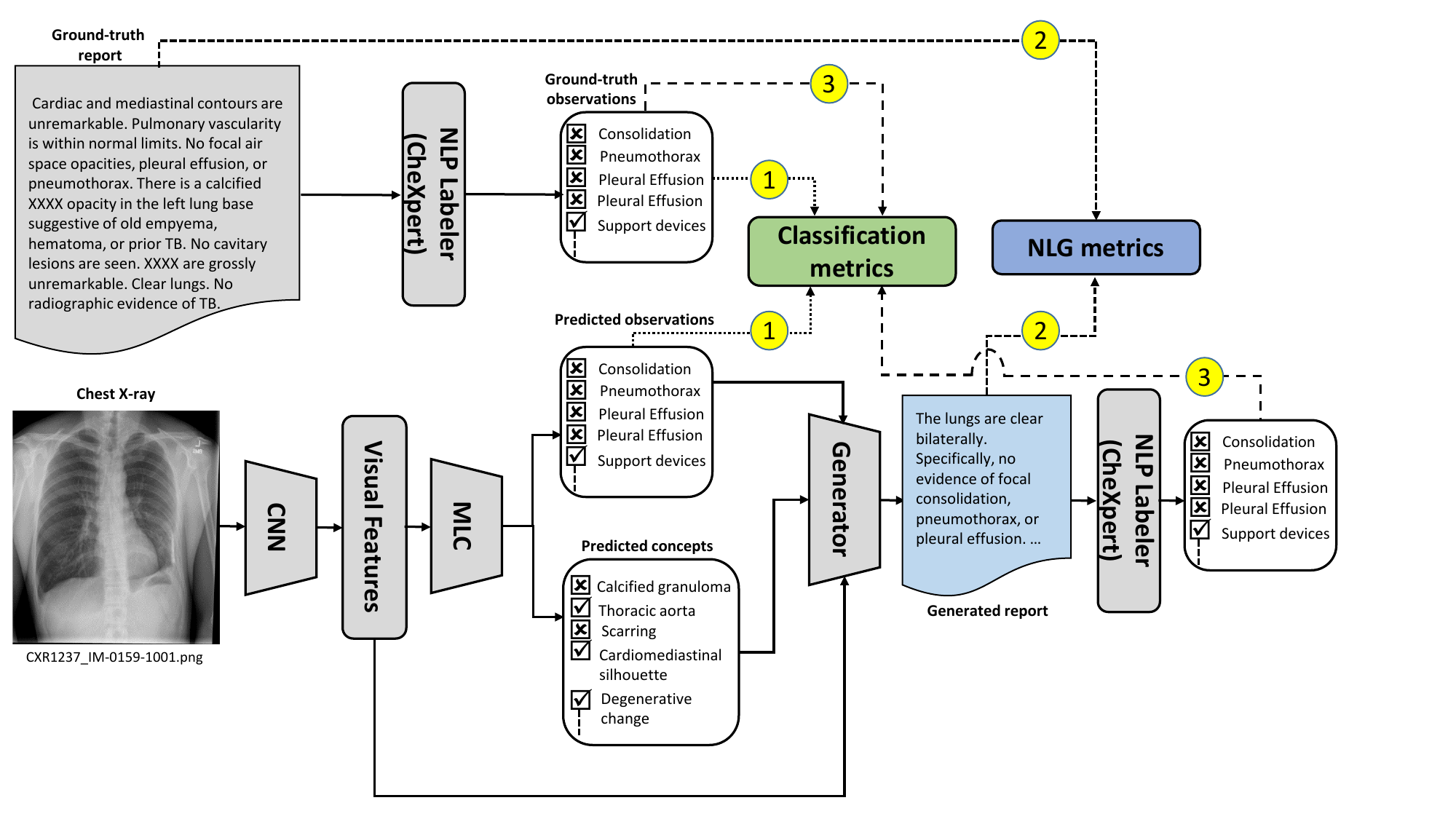}
    \caption{Detailed architecture of clinical context-aware radiology report generation.}
    \label{fig:evaluation_nlg_plus_diagnostic}
\end{figure}

To evaluate radiology reports from the clinical perspective, we propose a clinical context-aware radiology report generation. Figure~\ref{fig:evaluation_nlg_plus_diagnostic} shows the detailed architecture of a clinical context-aware radiology report generation method. In this method, we propose a three-step evaluation strategy based on the CheXpert labeler. Given the chest X-rays and radiology reports as a multi-modal data for radiology report generation, we first apply CheXpert labeler to extract ground-truth observations from the ground-truth radiology report. During testing stage, for an input chest X-ray, the multi-label classifier (MLC) predicts \emph{observations} and \emph{clinical tags}. The first step of evaluation is to compare \emph{ground-truth observations} with \emph{predicted observations} using standard \emph{classification metrics}, such as \emph{precision}, \emph{recall}, and \emph{f1-score}, in a multi-label setting. Given the availability of \emph{ground-truth tags} in the dataset, we can also compare the \emph{ground-truth tags} with the \emph{predicted tags} using \emph{classification metrics}, letting us know how well the MLC is performing in predicting correct clinical concepts, which are important for the generation stage. The second set of evaluation is to compare the \emph{ground-truth report} with the \emph{generated report} using standard \emph{natural language generation} (NLG) metrics, such as BLEU, METEOR, ROUGE, and CIDEr. In order to validate whether the generator is able to successfully use the predicted observations in the generated report, we use a third set of evaluation by comparing the \emph{ground-truth observations} with the \emph{observations} we get after applying the CheXpert labeler on the generated radiology report. This set of evaluation helps us to understand how well the generator is performing as well as analysing error at the classification stage and the generation stage.  

The proposed clinical context-aware radiology report generation system not only overcomes the limitation of non-differential nature of the CheXpert labeler, it also provides multi-stage evaluation strategy, which helps to utilise the complementary nature of NLG metrics and classification metrics. On one hand, the NLG metrics can highlight how well generated reports are fluent, grammatically correct, and coherent. On the other hand, the classification metrics can evaluate generated reports highlighting their clinical value. Also, having a multi-stage evaluation strategy helps us to perform error analysis at each levels of the radiology report generation pipeline, providing further insights for improvement at each stage, in turn providing room for improving the overall system. 

\subsection{A walk-through example}
\begin{figure}
    \centering
    \includegraphics[scale=0.4]{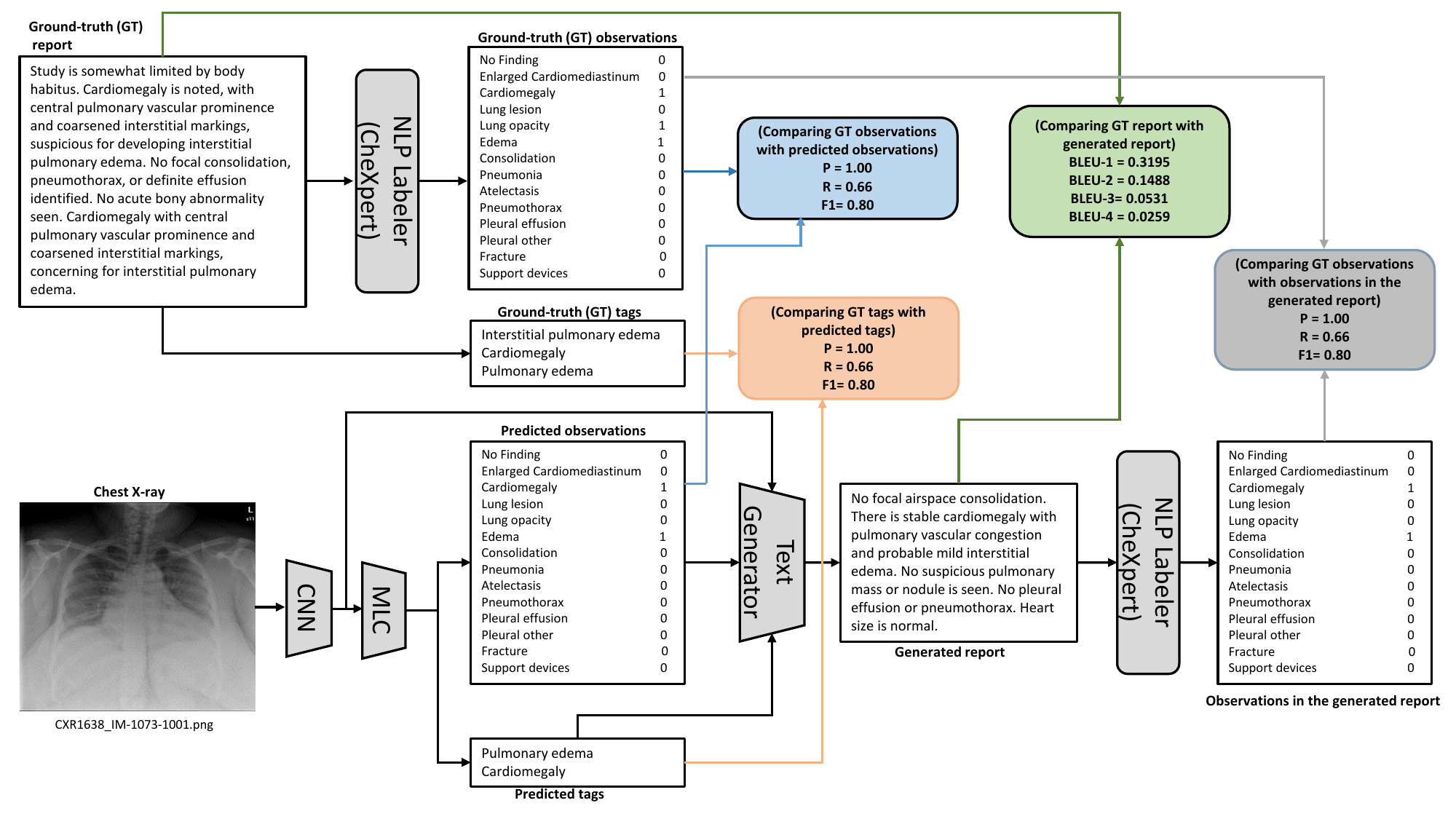}
    \caption{A walk-through example of robust evaluation for radiology report generation. CNN: Convolutional Neural Network; MLC: Multi-label Classifier; P: Precision; R: Recall: F1: F1-score.}
    \label{fig:a_walkthrough_example}
\end{figure}

To provide a better understanding of the clinical context-aware radiology report generation, in Figure~\ref{fig:a_walkthrough_example}, we provide a walk-through example with both the NLG metrics and classification metrics scores. Here, we assume that we have been provided with a dataset of chest X-rays accompanied by the radiology reports and clinical concepts or tags. Here we go through an example during the inference stage of radiology report generation. For a given chest X-ray, we first apply a CNN model to obtain a visual representation of the input image. These visual representations are passed to the Multi-label Classifier (MLC), which predicts labels for fourteen thoracic labels as well as for the tags present in the image. The visual representation of the image, embeddings of the predicted observations and tags are passed on to the text generator, which generates the radiology report. 

In order to evaluate the MLC, we apply the CheXpert labeler to the ground-truth report to obtain the ground-truth observations. We compare the ground-truth observations with the predicted observations by the MLC. In this example, we find that there are three thoracic diseases present in the study, namely, \emph{cardiomegaly}, \emph{lung opacity}, and \emph{edema}. However, the MLC is able to correctly predict only two of them, and missing \emph{lung opacity}. Given the ground-truth tags which are provided along with radiology reports, we compare the ground-truth tags with the predicted tags. Here, the MLC is able to correctly predict two of the clinical tags, ignoring the third one, namely, \emph{interstitial pulmonary edema}. To evaluate how fluent and coherent the generated report is, we apply NLG metrics comparing the ground-truth report with the generated report. In order to check how well the text generator is able to make use of the predicted observations and tags, we further apply the CheXpert labeler to the generated report to obtain the observations. We then compared the ground-truth observations with the observations present in the generated report in terms of standard classification metrics. Overall, this multi-step evaluation provides a robust measure of the radiology report generation system. Also, using both the NLG metrics and classification metrics helps in providing insight about how fluent, coherent, and diagnostically accurate the generated reports are when compared to the ground-truth reports. 

%%%%%%%%%%%%%%%%%%%%%%%%%%%%%%%%%%%%%%%%%%%%%%%%%%%%%%%%%%%%%%%%%%%%%%%%%
%%%%%%%%%%%%%%%%%%%%%%%%%%%%%%%%%%%%%%%%%%%%%%%%%%%%%%%%%%%%%%%%%%%%%%%%%
\section{Experiments}

\subsection{Datasets}
We use the Indiana University Chest X-ray collection (IU-CXR) dataset~\citep{Demner-Fushman:2015:Preparating_a_collection}, containing chest X-rays and their accompanied radiology reports. The IU-CXR dataset has 7,470 chest X-rays and 3,955 radiology reports. Each radiology report has sections including \emph{Comparison}, \emph{Indication}, \emph{Findings}, and \emph{Impression}. As the goal of this study is to check the effectiveness of the Transformer model as a decoder for radiology report generation, we do concatenation of the findings and impression section as the target radiology report for the model to be generated. We pre-process text data by doing tokenisation, converting to lower case, and filtering tokens having frequency less than three.

\subsection{Model training}
We use PyTorch~\citep{PyTorch}, a deep learning framework for implementing the proposed model in Python language. We use ResNet models~\citep{He:2016:Deep_residual_learning}, namely ResNet-18, ResNet-50, ResNet-101, and ResNet-152 as CNN encoder in our experiments. The word embedding size of LSTM is set to 512 in all cases.  We set the initial learning rate to 0.0001 and Adam optimiser~\citep{Kingma:2015:Adam} is used for updating the parameters of the model during training. We set $\beta_1$ to $0.9$ and $\beta_2$ to $0.999$ for the Adam optimiser. During training, the optimal model is saved based on the best possible BLEU score and termination is done by early stopping. To observe the effect of varying the number of hidden units in an LSTM, the number of hidden units are set to either 256, 512, or 1024 units. Also, for the CNN+Transformer network, experiments were performed using 1 head, 2 heads, or 3 heads. Also, we set the number of layers to either 3, 5, or 7 to see the effect of increasing the number of transformer layers. For all the experiments, we set the batch size to $8$ to manage the memory usage. We train our model under different settings for $30$ epochs and results for the test set are reported based on the best model configuration. To avoid overfitting of the model, we use \emph{dropout} as a regularisation term, with a value of $0.5$. During inference, we set the beam size to $3$ for radiology report generation. In order to evaluate the effectiveness of the model, we use standard NLG metrics, namely, BLEU, CIDEr, METEOR, and ROUGE.

\subsection{Results}
In order to investigate whether the \emph{Transformer} model is able to generate better quality radiology reports from chest X-rays, we conducted experiments with different set of parameters. To have a fair comparison, we keep the hyper-parameters of each model the same but make changes to either the encoder network or the decoder network. 

Table~\ref{tab:varying_encoder} shows results for radiology report generation for the CNN+LSTM framework with increasing complexity of the CNN model as an encoder. From the results, it is evident that with increase in the number of convolutional layers in the CNN model, the performance of radiology report generation increases. This reflects that deep and complex CNN models are better in terms of encoding chest X-rays. However, we find that the IU-CXR dataset is small, resulting in overfitting by the CNN model. We further investigate the effect of fine-tuning the CNN encoder in the CNN+LSTM framework as shown in Table~\ref{tab:finetuning_encoder}. The experimental results shows improvement in radiology report generation by fine-tuning the CNN encoder. We also investigate the effect of increasing the number of hidden units in the LSTM decoder for the CNN+LSTM framework. The experimental results in Table~\ref{tab:varying_hidden_units_LSTM_decoder} shows that increasing the number of hidden units in the LSTM decoder helps to improve radiology report generation. 

\begin{table}
  \centering
  \caption{Experimental results for varying encoder in the CNN + LSTM architecture.}
  \label{tab:varying_encoder}
  \begin{tabular}{lccccccc}
  \toprule
  Encoder & BLEU-1 & BLEU-2 & BLEU-3 & BLEU-4 & METEOR & ROUGE & CIDEr \\
 \midrule
    ResNet-18 &  33.72 & 21.44 & 14.94 & 10.98 & 14.71 & 28.16 & 43.74 \\ 
    ResNet-50 & 33.85 & 21.16 & 14.50 & 10.50 & 14.38 & 27.06 & 32.84 \\ 
    \rowcolor{Gray2} ResNet-101 & 36.80 & 23.20 & 16.02 & 11.49 & 15.78 & 29.31 & 40.67 \\ 
    ResNet-152 & 34.17 & 21.26 & 14.51 & 10.31 & 14.62 & 27.20 & 35.20 \\ 
 \bottomrule
 \end{tabular}
\end{table}

\begin{table}
  \centering
  \caption{Experimental results for fine-tuning encoder in the CNN + LSTM architecture.}
  \label{tab:finetuning_encoder}
  \begin{tabular}{lccccccc}
  \toprule
 Encoder & BLEU-1 & BLEU-2 & BLEU-3 & BLEU-4 & METEOR & ROUGE & CIDEr \\
 \midrule
    ResNet-18 & 32.86 & 20.83 & 14.36 & 10.42 & 14.40 & 27.19 & 34.88 \\  
    ResNet-50 & 33.57 & 21.32 & 14.86 & 10.87 & 14.81 & 28.13 & 36.93 \\ 
    \rowcolor{Gray2} ResNet-101 & 33.95 & 22.10 & 15.75 & 11.86 & 15.04 & 29.06 & 46.30 \\ 
    ResNet-152 & 31.65 & 19.91 & 13.64 & 9.80 & 14.08 & 26.84 & 32.30 \\ 
 \bottomrule
 \end{tabular}
\end{table}

\begin{table}
  \centering
  \caption{Experimental results for varying hidden units in the LSTM decoder in the CNN + LSTM architecture.}
  \label{tab:varying_hidden_units_LSTM_decoder}
  \begin{tabular}{lcccccccc}
  \toprule
 Encoder & \#HU & BLEU-1 & BLEU-2 & BLEU-3 & BLEU-4 & METEOR & ROUGE & CIDEr \\
 \midrule
    \multirow{3}{*}{ResNet-18} & 256 & 33.50 & 21.05 & 14.52 & 10.56 & 14.54 & 27.22 & 36.43 \\
                               & 512 & 33.72 & 21.44 & 14.94 & 10.98 & 14.71 & 28.16 & 43.74 \\
                               & 1024 & 34.35 & 21.19 & 14.65 & 10.74 & 14.45 & 27.00 & 34.18 \\ \hline 
    \multirow{3}{*}{ResNet-50} & 256 &  33.62 & 20.46 & 13.69 & 9.60 & 14.19 & 26.02 & 29.03 \\ 
                               & 512 & 33.85 & 21.16 & 14.50 & 10.50 & 14.38 & 27.06 & 32.84 \\
                               & 1024 & 35.09 & 21.77 & 14.88 & 10.78 & 14.73 & 26.62 & 33.41 \\ \hline 
    \multirow{3}{*}{ResNet-101} & 256 &  34.13 & 21.28 & 14.34 & 10.07 & 14.59 & 27.25 & 31.61 \\ 
                               & 512 & 36.11 & 23.20 & 16.02 & 11.49 & 15.78 & 29.31 & 40.67 \\
                               & 1024 & 34.55 & 21.17 & 14.33 & 10.27 & 14.44 & 25.91 & 22.28 \\ \hline 
    \multirow{3}{*}{ResNet-152} & 256 & 35.74 & 22.42 & 15.34 & 10.84 & 15.30 & 28.33 & 35.40  \\ 
                               & 512 & 34.17 & 21.26 & 14.51 & 10.31 & 14.62 & 27.20 & 35.20 \\
            \rowcolor{Gray2}  & 1024 & 36.80 & 23.28 & 16.46 & 12.31 & 15.48 & 28.63 & 42.55 \\ 
 \bottomrule
 \end{tabular}
\end{table}

In order to investigate the effect of varying parameters in the CNN+Transformer framework, we first check the effect of increasing the number of convolutional layers in the CNN model. The experimental results in Table~\ref{tab:varying_encoder_transformer} shows that better CNN model used as an encoder provides better results for radiology report generation. To validate the effect of fine-tuning the CNN model, we fine-tune the last layer of the CNN model. The experimental results in Table~\ref{tab:finetuning_encoder_transformer} shows the positive effect of fine-tuning the encoder for radiology report generation. To validate the hypothesis of increasing complexity of the Transformer model as the decoder, we vary the number of heads and number of layers in the Transformer model in the CNN+Transformer framework. The experimental results in Table~\ref{tab:varying_Transformer_decoder_with_resnet18} shows that increasing number of heads and increasing the number of layers in the transformer decoder improves radiology report generation. It means that more decoder layers can make the model achieve better results. However, with more layers, the decoder will have more parameters, making the model more complex. 

\begin{table}
  \centering
  \caption{Experimental results for varying encoder in the CNN + Transformer architecture.}
  \label{tab:varying_encoder_transformer}
  \begin{tabular}{llccccccc}
  \toprule
 Encoder & BLEU-1 & BLEU-2 & BLEU-3 & BLEU-4 & METEOR & ROUGE & CIDEr \\
 \midrule
    ResNet-18 &  31.98 & 19.68 & 13.06 & 9.09 & 13.58 & 25.89 & 29.59 \\ 
    ResNet-50 &  29.64 & 18.14 & 11.62 & 7.63 & 13.37 & 25.60 & 20.75 \\ 
    \rowcolor{Gray2} ResNet-101 &  31.19 & 19.66 & 13.13 & 9.10 & 13.66 & 26.47 & 28.97 \\ 
    ResNet-152 &  29.31 & 18.23 & 11.85 & 7.69 & 13.43 & 26.23 & 22.51 \\ 
 \bottomrule
 \end{tabular}
\end{table}

\begin{table}
  \centering
  \caption{Experimental results for fine-tuning encoder in the CNN + Transformer architecture.}
  \label{tab:finetuning_encoder_transformer}
  \begin{tabular}{lccccccc}
  \toprule
 Encoder & BLEU-1 & BLEU-2 & BLEU-3 & BLEU-4 & METEOR & ROUGE & CIDEr \\
 \midrule
    ResNet-18 & 30.69 & 19.42 & 13.23 & 9.41 & 13.53 & 26.56 & 30.81 \\ 
    ResNet-50 &  29.61 & 17.83 & 11.55 & 7.74 & 13.30 & 25.94 & 19.42 \\
   \rowcolor{Gray2} ResNet-101 &  31.36 & 19.76 & 13.34 & 9.52 & 14.17 & 27.73 & 32.90 \\ 
    ResNet-152 &  30.51 & 18.85 & 12.66 & 8.93 & 13.69 & 26.73 & 30.26 \\ 
 \bottomrule
 \end{tabular}
\end{table}

\begin{table}
  \centering
  \caption{Experimental results for varying Transformer decoder in the CNN+Transformer architecture. We fix the ResNet-18 model as an encoder for all results in this table.}
  \label{tab:varying_Transformer_decoder_with_resnet18}
  \begin{tabular}{ccccccccc}
  \toprule
 Heads & Layers & BLEU-1 & BLEU-2 & BLEU-3 & BLEU-4 & METEOR & ROUGE & CIDEr \\
 \midrule
    \multirow{3}{*}{1} & 3 &  31.98 & 19.68 & 13.06 & 9.09 & 13.58 & 25.89 & 29.59\\
                       & 5 &  32.08 & 20.11 & 13.62 & 9.55 & 14.06 & 27.07 & 33.24 \\
                       & 7 &  31.70 & 19.50 & 13.01 & 9.02 & 13.66 & 26.25 & 29.86 \\ \hline 
    \multirow{3}{*}{2} & 3 &  31.22 & 19.04 & 12.47 & 8.52 & 13.41 & 25.68 & 27.47 \\
                       & 5 &  32.28 & 20.01 & 13.37 & 9.32 & 13.99 & 26.73 & 30.43 \\
                       & 7 &  31.19 & 19.43 & 12.96 & 8.96 & 13.65 & 26.38 & 28.50 \\ \hline 
    \rowcolor{Gray2} 
    \multirow{3}{*}{3} & 3 &  32.52 & 20.11 & 13.54 & 9.46 & 14.01 & 26.88 & 29.38 \\
                        & 5 &  31.71 & 19.48 & 12.96 & 9.09 & 13.44 & 25.79 & 30.08 \\
                       & 7 &  30.38 & 19.06 & 12.87 & 9.01 & 13.71 & 26.67 & 26.40 \\ \hline 
 \bottomrule
 \end{tabular}
\end{table}

\begin{table}
  \centering
  \caption{Experimental results for varying Transformer decoder in the CNN+Transformer architecture. We fix the ResNet-101 model with fine-tuning as an encoder for all results in this table.}
  \label{tab:varying_Transformer_decoder_with_resnet101}
  \begin{tabular}{ccccccccc}
  \toprule
 Heads & Layers & BLEU-1 & BLEU-2 & BLEU-3 & BLEU-4 & METEOR & ROUGE & CIDEr \\
 \midrule
    \multirow{3}{*}{1} & 3 &  33.36 & 21.76 & 14.34 & 10.52 & 14.17 & 27.73 & 32.90 \\
                       & 5 &  33.95 & 21.98 & 14.65 & 10.95 & 14.06 & 27.25 & 41.46 \\
                       & 7 &  32.98 & 21.04 & 14.72 & 10.02 & 13.85 & 26.63 & 29.40 \\ \hline 
    \multirow{3}{*}{2} & 3 &  35.39 & 22.89 & 16.10 & 11.93 & 15.50 & 29.20 & 36.14 \\
                       & 5 & 36.10 & 23.64 & 16.81 & 12.54 & 15.96 & 30.24 & 37.38  \\
                       & 7 & 36.17 & 23.49 & 16.79 & 12.76 & 15.54 & 29.33 & 37.94  \\ \hline 
    \multirow{3}{*}{3} & 3 & 36.62 & 23.45 & 16.20 & 13.20 & 15.99 & 29.72 & 38.87 \\ 
               \rowcolor{Gray2} 3  & 5 & 36.92 & 23.78 & 16.62 & 13.70 & 16.40 & 31.08 & 43.38  \\
                       & 7 & 36.58 & 23.30 & 16.15 & 13.00 & 15.36 & 29.53 & 38.77  \\ \hline 
 \bottomrule
 \end{tabular}
\end{table}

After extensive sets of experiments by varying parameters on both the encoder and decoder side, we use the optimal parameters for both the CNN+LSTM and the CNN+Transformer models to report results on the radiology report generation. Table~\ref{tab:results_transformer_captioning} shows results of radiology report generation for the two models. The experimental results shows that using the transformer as a decoder, provides superior results compared to the LSTM network. This shows the effectiveness of using better language models in improving the radiology report generation. 

\begin{table}
  \centering
  \caption{Performance comparison of the \{CNN + LSTM\}$^*$ model with the \{CNN + Transformer\}$^*$ model for radiology report generation on the IU-CXR dataset. }
  \label{tab:results_transformer_captioning}
  \begin{tabular}{lccccccc}
  \toprule
 Model & BLEU-1 & BLEU-2 & BLEU-3 & BLEU-4 & METEOR & ROUGE & CIDEr \\
 \midrule
 \{CNN+LSTM\}$^*$ & 36.80 & 23.28 & 16.46 & 12.31 & 15.48 & 28.63 & 42.55 \\ 
 \rowcolor{Gray2}
\{CNN+Transformer\}$^*$ & 36.92 & 23.78 & 16.62 & 13.70 & 16.40 & 31.08 & 43.38 \\
 \bottomrule
 \end{tabular}
 \begin{minipage}{14.5cm}
 $^*$ denotes optimal hyperparameters settings for both the encoder and decoder to get best configuration for both of the models. We use best configurations based on the parameter analysis and choosing best results for both models.
 \end{minipage}
\end{table}

In order to investigate the time complexity of varying decoders, we note time taken to train the CNN+LSTM and the CNN+Transformer model for radiology report generation on the IU-CXR dataset. We execute our training scripts on a single Nvidia Tesla Volta V100 GPU, keeping all the other parameters same. We use Adam optimiser with learning rate of $0.0001$, $\beta_1$ of $0.9$, $\beta_2$ of $0.999$, batch size of $8$, and keeping the ResNet-18 as the encoder. For the CNN+LSTM model, increasing the number of LSTM hidden units in the decoder increases the training time, which varies linearly with an increase in decoder complexity. The similar trend is shown for the CNN+Transformer model, where increasing number of heads and number of decoding layers increases the training time. Comparing the time taken for training the CNN+LSTM with the CNN+Transformer model, on the IU-CXR dataset, we find that the training time for CNN+Transformer is significantly lower than the CNN+LSTM model. This validates the hypothesis that fully-attention based networks such as the transformer are faster to train due to their reliance on the non-recurrence mechanism and can be trained in parallel.

%%%%%%%%%%%%%%%%%%%%%%%%%%%%%%%%%%%%%%%%%%%%%%%%%%%%%%%%%%%%%%%%%%%%%
%%%%%%%%%%%%%%%%%%%%%%%%%%%%%%%%%%%%%%%%%%%%%%%%%%%%%%%%%%%%%%%%%%%%%
\section{Discussion}
In this study, based on our extensive experiments, we recommend the use of state-of-the-art language models as decoders for the radiology report generation. Traditional RNN networks as decoders have limitations in terms of not capturing long-range context, which is important for generating long and coherent radiology report. On the other hand, the transformer model not only helps to capture context based on visual features, thoracic observations, clinical concepts, and textual radiology reports, it also uses far fewer parameters and providing faster execution. The results of radiology report generation are promising but still there are many open-ended challenges which need to be solved. Given that the IU-CXR dataset is of small size, we believe that the transformer model is a complex model, which need large-scale dataset to improve generalisation. However, in our experiments as described in the previous section, we find that our CNN+Transformer model starts overfitting when the model becomes too complex due to the increased number of heads and stacking multiple decoding layers. We believe that state-of-the-art language models such as the transformer can really shine when large-scale dataset consisting of chest X-rays with their accompanied radiology reports are used to train the model. 

\begin{figure}
    \centering
    \includegraphics[scale=0.4]{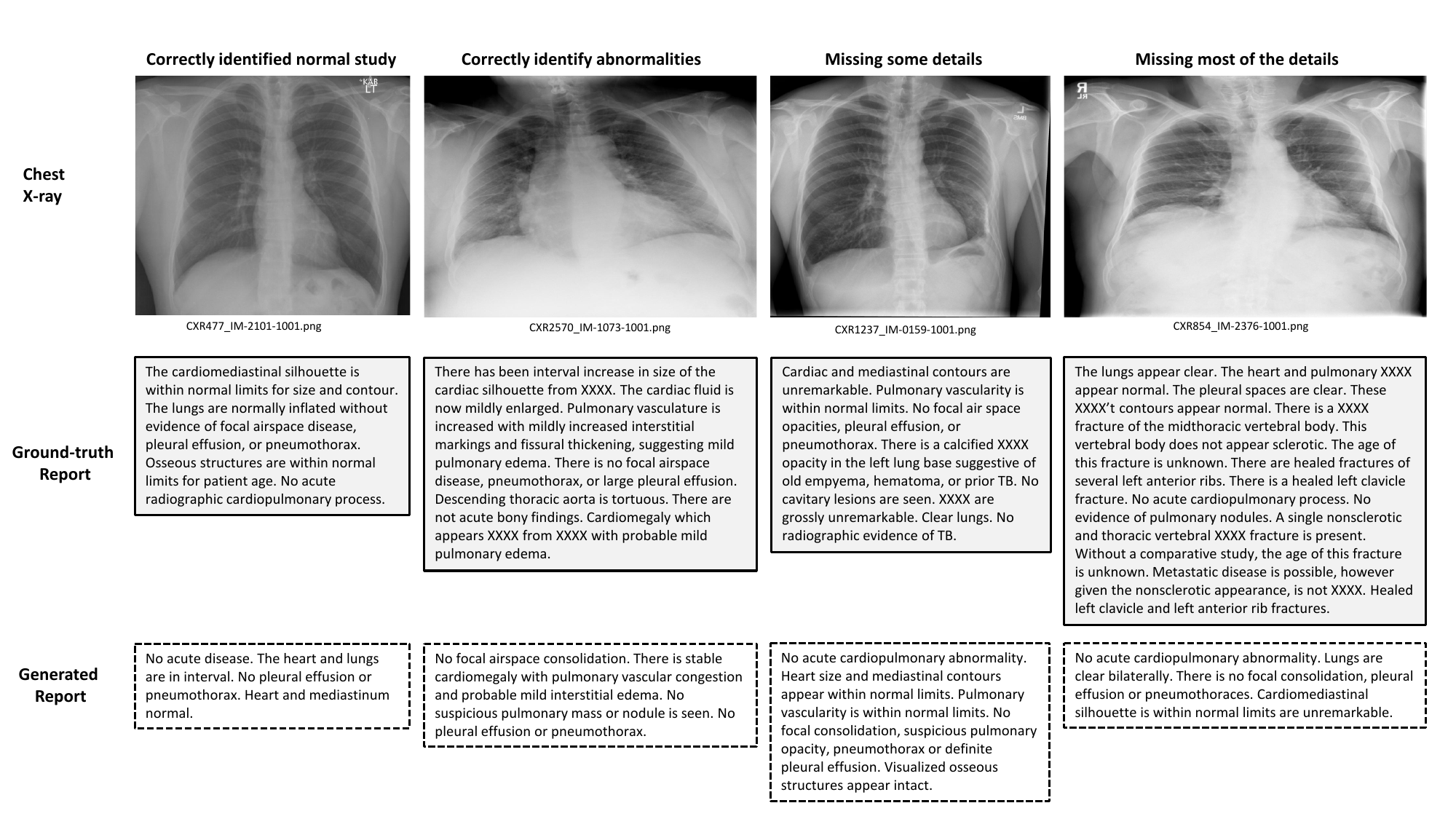}
    \caption{Selected sample cases for radiology report generation with increasing difficulty. Bold words represent statements that produce the same meaning as in the ground-truth report.}
    \label{fig:four_case_analysis}
\end{figure}

Figure~\ref{fig:four_case_analysis} highlight four selected cases for radiology report generation. The first case shows the model's ability to correctly identify a normal study by highlighting salient normal regions in the chest X-ray. Given that the number of normal studies outnumber the abnormal studies in most medical imaging datasets, the first case is an easy problem when the language model can easily learn from majority of the text data. In the second case, model can correctly identify abnormalities such as \emph{cardiomegaly}, \emph{pulmonary vascular congestion}, and \emph{interstitial edema}. However, the model is unable to generate sentences such as ``\emph{there has been interval increase in size of cardiac silhouette from XXXX}". Based on our analysis, we find that radiologists have previous studies for comparison and this is a similar case, where radiologist compare current study with the previous one having date XXXX. Because of not utilising background information for radiology report generation, our model is unable to generated such sentences. We also find that certain words such as \emph{stable} are temporal in nature. Most of the studies, including our current study, do not have access to longitudinal data, it is hard to apply temporal learning in radiology report generation methods. In the third case, the model is able to correctly highlight normal regions, but misses key findings of \emph{calcified XXXX opacity}. We find this is due to the lack of context for the need of current radiographic study. In the \emph{indication} section, physician writes why the particular radiology study is asked for. This background context can guide radiologists to look specifically to answer the clinical question. On checking the indication section in the ground-truth report of the third case, we find sentence ``XXXX-year-old male with positive PPD", which clearly indicates that radiologist is investigating whether radiology study shows any symptoms of tuberculosis (TB). This is evident from the sentence ``No radiographic evidence of active TB" in the ground-truth report. The fourth example highlights one of the challenging case, given the nature of the study and usage of words in the report. Given cases related to \emph{fractures} are \emph{rare} in the dataset, it is challenging for language model to learn and generate such sort of examples.  

Due to the HIPAA compliance, the medical imaging datasets are made publicly available after the de-identification of both the medical images and textual radiology reports. In medical images, generally in DICOM format, the personal health information (PHI) is generally located on the top left of the DICOM header. Removing this PHI from DICOM headers and converting file to other formats, usually does not degrade the usability of the data, except decreasing the image resolution. However, when applying de-identification algorithms on the textual radiology reports, many extra tokens are removed, reducing the usability of radiology reports. In the IU-CXR dataset, some reports consists of erroneous words such as ``XXXX" or ``X-XXXX", which are important keywords from diagnostic viewpoint, but are erroneously removed during the de-identification process. This noisy information has an adverse effect on the performance of language models. We encourage researchers to carefully curate datasets and tune de-identification algorithms so that it does not remove non-PHI information from the dataset. 

In this study, we focus on the generation-based approach for radiology report generation. In the past, research has been towards combining the template-based approach with the generation-based approach~\citep{Li:2019:Knowledge_driven_encode_retrieve}. Given different approaches of text generation, namely, template-based, retrieval-based, and generation-based have both pros and cons, in the future, we aim to combine these approaches to further improve the radiology report generation task. While research has shown promising results for generating radiology reports from medical images, the efficacy of developed models in actual radiology practice is still not explored. One of the future directions is towards validating radiology report generation models in clinical practice and to objectively measure its performance in terms of improving clinical workflow, reducing diagnostic errors, and augmenting radiologists. 

%%%%%%%%%%%%%%%%%%%%%%%%%%%%%%%%%%%%%%%%%%%%%%%%%%%%%%%%%%%%%%%%%%%%%
%%%%%%%%%%%%%%%%%%%%%%%%%%%%%%%%%%%%%%%%%%%%%%%%%%%%%%%%%%%%%%%%%%%%%
%%%%%%%%%%%%%%%%%%%%%%%%%%%%%%%%%%%%%%%%%%%%%%%%%%%%%%%%%%%%%%%%%%%%%
%%%%%%%%%%%%%%%%%%%%%%%%%%%%%%%%%%%%%%%%%%%%%%%%%%%%%%%%%%%%%%%%%%%%%
\section{Conclusion}
In this paper, we investigate the transformer model for the radiology report generation from medical images. Instead of using LSTM network as a standard decoder for radiology report generation, we use transformer as a decoder due to its inherent properties of being faster in execution, can be trained in parallel, less memory intensive, and captures global dependencies in multi-modal data. We did extensive experiments to see the effect of various parameters under different settings and bring insight based on results. The experimental results in terms of standard language generation metrics shows the superiority of the fully-attention based decoder compared to the recurrence based decoder. The proposed methodology has potential in improving clinical workflow, improving patient care, augmenting radiologists by providing second opinion, and stratifying cases that need urgent attention.

%%%%%%%%%%%%%%%%%%%%%%%%%%%%%%%%%%%%%%%%%%%%%%%%%%%%%%%%%%%%%%%
%%%%%%%%%%%%%%%%%%%%%%%%%%%%%%%%%%%%%%%%%%%%%%%%%%%%%%%%%%%%%%%
\section*{Declaration of competing interest}
The author(s) declare no conflicts of interest.

%%%%%%%%%%%%%%%%%%%%%%%%%%%%%%%%%%%%%%%%%%%%%%%%%%%%%%%%%%%%%
%%%%%%%%%%%%%%%%%%%%%%%%%%%%%%%%%%%%%%%%%%%%%%%%%%%%%%%%%%%%%

% Acknowledgements and Disclosure of Funding should go at the end, before appendices and references
% Manual newpage inserted to improve layout of sample file - not
% needed in general before appendices/bibliography.

%\newpage

\vskip 0.2in
\bibliography{sample}

\end{document}